%% file: main.tex
\documentclass[10pt,twocolumn,letterpaper]{article}

\usepackage[pagenumbers]{cvpr} 

\makeatletter
\@namedef{ver@everyshi.sty}{}
\makeatother

\input{Preamble/preamble.tex}
\input{Preamble/acronyms.tex}

\input{Preamble/symbols.tex}

\definecolor{cvprblue}{rgb}{0.21,0.49,0.74}
\usepackage[breaklinks,colorlinks,citecolor=cvprblue]{hyperref}

\graphicspath{{Images/}}

\def\challink{\url{https://codalab.lisn.upsaclay.fr/competitions/17161}}

\newcommand{\team}[1]{%
\ifcase#1
\textbf{Baseline}\xspace
\or PICO-MR\xspace
\or Anonymous\xspace
\or RGA-Robot\xspace
\or EVP++\xspace
\or Anonymous\xspace
\or 3DCreators\xspace
\or visioniitd\xspace
\or Anonymous\xspace
\or HIT-AIIA\xspace
\or FRDC-SH\xspace
\or Anonymous\xspace
\or Anonymous\xspace
\or Anonymous\xspace
\or Anonymous\xspace
\or hyc123\xspace
\or ReadingLS\xspace
\or Anonymous\xspace
\or Anonymous\xspace
\or Elder Lab\xspace
\else
Invalid number%
\fi
}

\newcommand{\train}[1]{%
\ifcase#1
S\xspace
\or $\dagger$D*\xspace
\or ?\xspace
\or $\dagger$S\xspace
\or $\dagger$D\xspace
\or ?\xspace
\or $\dagger$D\xspace
\or D\xspace
\or ?\xspace
\or $\dagger$D\xspace
\or $\dagger$D\xspace
\or ?\xspace
\or ?\xspace
\or ?\xspace
\or ?\xspace
\or D\xspace
\or $\dagger$MD*\xspace
\or ?\xspace
\or ?\xspace
\or D\xspace
\else
Invalid number%
\fi
}

\newcommand{\teamsec}[1]{Team #1: \team{#1} -- \train{#1}}

\usepackage[capitalize]{cleveref}
\crefname{section}{Sec.}{Secs.}
\Crefname{section}{Section}{Sections}
\Crefname{table}{Table}{Tables}
\crefname{table}{Tab.}{Tabs.}

\def\cvprPaperID{1} 
\def\confName{CVPR}
\def\confYear{2024}

\makeatletter
\let\@fnsymbol\@arabic
\makeatother

\begin{document}

\title{The Third Monocular Depth Estimation Challenge}

\author{%
Jaime Spencer\thanks{Independent} \and
Fabio Tosi\thanks{University of Bologna} \and
Matteo Poggi\footnotemark[2] \and
Ripudaman Singh Arora\thanks{Blue River Technology} \and
Chris Russell\thanks{Oxford Internet Institute} \and
Simon Hadfield\thanks{University of Surrey} \and
Richard Bowden\footnotemark[5] \and
GuangYuan Zhou\thanks{ByteDance} \and
ZhengXin Li\thanks{University of Chinese Academy of Science} \and
Qiang Rao\footnotemark[6] \and
YiPing Bao\footnotemark[6] \and
Xiao Liu\footnotemark[6] \and
Dohyeong Kim\thanks{RGA Inc.} \and
Jinseong Kim\footnotemark[8] \and
Myunghyun Kim\footnotemark[8] \and
Mykola Lavreniuk\thanks{Space Research Institute NASU-SSAU, Kyiv, Ukraine} \and
Rui Li\thanks{Northwestern Polytechnical University, Xi'an} \and
Qing Mao\footnotemark[10] \and
Jiang Wu\footnotemark[10] \and
Yu Zhu\footnotemark[10] \and
Jinqiu Sun\footnotemark[10] \and
Yanning Zhang\footnotemark[10] \and
Suraj Patni\thanks{Indian Institute of Technology, Delhi} \and
Aradhye Agarwal\footnotemark[11] \and
Chetan Arora\footnotemark[11] \and
Pihai Sun\thanks{Harbin Institute of Technology} \and
Kui Jiang\footnotemark[12] \and
Gang Wu\footnotemark[12] \and
Jian Liu\footnotemark[12] \and
Xianming Liu\footnotemark[12] \and
Junjun Jiang\footnotemark[12] \and
Xidan Zhang\thanks{Fujitsu} \and
Jianing Wei\footnotemark[13] \and
Fangjun Wang\footnotemark[13] \and 
Zhiming Tan\footnotemark[13] \and
Jiabao Wang\thanks{GuangXi University} \and
Albert Luginov\thanks{University of Reading} \and
Muhammad Shahzad\footnotemark[15] \and
\and
Seyed Hosseini\thanks{York University} \and
Aleksander Trajcevski\footnotemark[16] \and 
James H. Elder\footnotemark[16] 
}
\maketitle

\begin{abstract}
This paper discusses the results of the third edition of the Monocular Depth Estimation Challenge (MDEC). 
The challenge focuses on zero-shot generalization to the challenging SYNS-Patches dataset, featuring complex scenes in natural and indoor settings. 
As with the previous edition, methods can use any form of supervision, i.e. supervised or self-supervised. 
The challenge received a total of 19 submissions outperforming the baseline on the test set: 10 among them submitted a report describing their approach, highlighting a diffused use of foundational models such as Depth Anything at the core of their method. The challenge winners drastically improved 3D F-Score performance, from 17.51\% to 23.72\%.
\end{abstract}

\section{Introduction} \label{sec:intro}
\Ac{mde} aims at predicting the distance from the camera to the points of the scene depicted by the pixels in the captured image.
It is a highly ill-posed problem due to the absence of geometric priors usually available from multiple images. Nonetheless, deep learning has rapidly advanced this field and made it a reality, enabling results far beyond imagination.

For years, most proposed approaches have been tailored to training and testing in a single, defined domain -- e.g., automotive environments \cite{Geiger2013} or indoor settings \cite{Silberman2012} -- often ignoring their ability to generalize to unseen environments.
Purposely, the \ac{mdec} in the last years has encouraged the community to delve into this aspect, by proposing a new benchmark for evaluating \ac{mde} models on a set of complex environments, comprising natural, agricultural, urban, and indoor settings. The dataset comes with a validation and a testing split, without any possibility of training/fine-tuning over it thus forcing the models to generalize.

While the first edition of \ac{mdec}~\cite{Spencer2023} focused on benchmarking self-supervised approaches, the second \cite{spencer2023second} additionally opened the doors to supervised methods.
During the former, the participants outperformed the baseline~\cite{Spencer2022,Garg2016} in all image-based metrics (AbsRel, MAE, RMSE), but could not improve pointcloud reconstructions~\cite{Ornek2022} (F-Score). The latter, instead, brought new methods capable of outperforming the baseline on both aspects, establishing a new \ac{sota}. 
The third edition of \ac{mdec}, detailed in this paper, ran in conjunction with CVPR2024, following the successes of the second one by allowing submissions of methods exploiting any form of supervision, \eg supervised, self-supervised, or multi-task.

Following previous editions, the challenge was built around \acl{syns}~\cite{Adams2016,Spencer2022}.
This dataset was chosen because of the variegated diversity of environments it contains, including urban, residential, industrial, agricultural, natural, and indoor scenes. 
Furthermore, \acl{syns} contains dense high-quality \acs{lidar} ground-truth, which is very challenging to obtain in outdoor settings.
This allows for a benchmark that accurately reflects the real capabilities of each model, potentially free from biases.

While the second edition counted 8 teams outperforming the \ac{sota} baseline in either pointcloud- or image-based metrics, this year 19 submissions achieved this goal. Among these, 10 submitted a report introducing their approach, 7 of whose outperformed the winning team of the second edition. 
This demonstrates the increasing interest -- and efforts -- in \ac{mdec}.

In the remainder of the paper, we will provide an overview of each submission, analyze their results on \acl{syns}, and discuss potential future developments.

\section{Related Work} \label{sec:lit}

\heading{Supervised MDE}
Early monocular depth estimation (MDE) efforts utilized supervised learning, leveraging ground truth depth labels. Eigen et al.~\cite{Eigen2015} proposed a pioneering end-to-end convolutional neural network (CNN) for MDE, featuring a scale-invariant loss and a coarse-to-fine architecture. Subsequent advancements incorporated structured prediction models such as Conditional Random Fields (CRFs)~\cite{Liu2015,Weihao2022} and regression forests~\cite{Roy2016}. Deeper network architectures~\cite{Xian2018,Ranftl2020}, multi-scale fusion~\cite{Miangoleh2021}, and transformer-based encoders~\cite{Ranftl2021,Cheng2021,Bhat2023} further enhanced performance. Alternatively, certain methods framed depth estimation as a classification problem~\cite{Fu2018,Li2019,Bhat2021,Bhat2022}. Novel loss functions were also introduced, including gradient-based regression~\cite{Li2018,Wang2019b}, the berHu loss~\cite{Laina2016a}, an ordinal relationship loss~\cite{Chen2016}, and scale/shift invariance~\cite{Ranftl2020}.

\heading{Self-Supervised MDE}
To overcome the dependence on costly ground truth annotations, self-supervised methods were developed. Garg et al.~\cite{Garg2016}, for the first time, proposed an algorithm based on view synthesis and photometric consistency across stereo image pairs, the importance of which for was extensively analyzed by Poggi et al.~\cite{Poggi2021}. Godard et al.~\cite{Godard2017} introduced Monodepth, which incorporated differentiable bilinear interpolation~\cite{Jaderberg2015}, virtual stereo prediction, and a SSIM+\pnorm{1} reconstruction loss. Zhou et al.~\cite{Zhou2017} presented SfM-Learner, which required only monocular video supervision by replacing the known stereo transform with a pose estimation network.
Following the groundwork laid by these frameworks, subsequent efforts focused on refining the depth estimation accuracy by integrating feature-based reconstructions~\cite{Zhan2018,Spencer2020,Yu2020}, semantic segmentation~\cite{Ramirez2019}, adversarial losses~\cite{Aleotti2018}, proxy-depth representations~\cite{Klodt2018,Rui2018,Andraghetti2019,Watson2019,Tosi2019,Choi2021,Peng2021}, trinocular supervision~\cite{Poggi2018} and other constraints~\cite{Mahjourian2018,Wang2018,Bian2019}. Other works focused on improving depth estimates at object boundaries~\cite{Tosi2021,Talker2024}. Moreover, attention has also been given to challenging cases involving dynamic scenarios during the training phase, which pose difficulties in providing accurate supervision signals for such networks. This has been addressed, for example, by incorporating uncertainty estimates~\cite{Klodt2018,Yang2020a,Poggi2020}, motion masks~\cite{Gordon2019,Casser2019,Dai2020,Tosi2020}, optical flow~\cite{Yin2018,Ranjan2019,Luo2020}, or via the minimum reconstruction loss~\cite{Godard2019}.
Finally, several architectural innovations, including 3D (un)packing blocks~\cite{Guizilini2020}, position encoding~\cite{Bello2021}, transformer-based encoders~\cite{Zhao2022,Agarwal2023}, sub-pixel convolutions~\cite{Pillai2019}, progressive skip connections~\cite{Lyu2021}, and self-attention decoders~\cite{Johnston2020,Yan2021,Zhou2021}, allowed further improvements. Among them, lightweight models tailored for real-time applications with memory and runtime constraints have also been developed~\cite{Poggi2018b,Peluso2019,Wofk2019,Aleotti2020,Peluso2021,Cipolletta2021,Huynh2022}.

\heading{Generalization and ``In-the-Wild" MDE} 
Estimating depth in the wild refers to the challenging task of developing methods that can generalize to a wide range of unknown settings~\cite{Chen2016,Chen2020}. Early works in this area focused on predicting relative (ordinal) depth~\cite{Chen2016,Chen2020}. Nonetheless, the limited suitability of relative depth in many downstream contexts has driven researchers to explore affine-invariant depth estimation~\cite{Li2018,Yin2020}.
In the affine-invariant setting, depth is estimated up to an unknown global offset and scale, offering a compromise between ordinal and metric representations. Researchers have employed various strategies to achieve generalization, including leveraging annotations from large datasets to train monocular depth models~\cite{Ranftl2021,Ranftl2020,Yang2024}, including internet photo collections~\cite{Li2018,Yin2020}, as well as from automotive LiDAR~\cite{Geiger2013,Guizilini2020,Huang2019}, RGB-D/Kinect sensors~\cite{Silberman2012,Cho2021,Sturm2012}, structure-from-motion reconstructions~\cite{Li2018,Li2020}, optical flow/disparity estimation~\cite{Xian2018,Ranftl2020}, and crowd-sourced annotations~\cite{Chen2016}.  However, the varying accuracy of these annotations may have impacted model performance, and acquiring new data sources remains a challenge, motivating the exploration of self-supervised approaches~\cite{Yin2021,Zhang2022}. 
For instance, KBR(++)~\cite{Spencer2023c,Spencer2024} leverage large-scale self-supervision from curated internet videos.
The transition from CNNs to vision transformers has further boosted performance in this domain, as demonstrated by DPT (MiDaS v3)~\cite{Ranftl2021} and Omnidata~\cite{Eftekhar2021}.
Furthermore, a few works like Metric3D~\cite{Yin2023} and ZeroDepth~\cite{Guizilini2023} revisited the depth estimation by explicitly feeding camera intrinsics as additional input. A notable recent trend involves training generative models, especially diffusion models~\cite{Song2020, Ho2020, Fu2024} for monocular depth estimation~\cite{Ji2023, Duan2023, Saxena2023, Saxena2023b, Ke2024}.


\addfig*[t]{distrib}{data:distrib}

\heading{Adverse Weather and Transparent/Specular Surfaces}
Existing monocular depth estimation networks have struggled under adverse weather conditions. Approaches have addressed low visibility~\cite{Spencer2020}, employed day-night branches using GANs~\cite{Vankadari2020,Zhao2022b}, utilized additional sensors~\cite{Gasperini2021}, or faced trade-offs~\cite{Vankadari2023}. Recently, md4all~\cite{Gasperini2023} enabled robust performance across conditions without compromising ideal setting performance.
Furthermore, estimating depth for transparent or mirror (ToM) surfaces posed a unique challenge~\cite{Ramirez2023b,Ramirez2024}. Costanzino et al.~\cite{Costanzino2023} is the only work dedicated to this, introducing novel datasets~\cite{Ramirez2022,Ramirez2023}. Their approach relied on segmentation maps or pre-trained networks, generating pseudo-labels by inpainting ToM objects and processing them with a pre-trained depth model~\cite{Ranftl2020}, enabling fine-tuning of existing networks to handle ToM surfaces.

\section{The Monocular Depth Estimation Challenge} \label{sec:meth}
The third edition of the \acl{mdec}\footnote{\challink} was organized on CodaLab~\cite{Pavao2022} as part of a CVPR2024 workshop.
The development phase lasted four weeks, using the \acl{syns} validation split.
During this phase, the leaderboard was public but the usernames of the participants were anonymized.
Each participant could see the results achieved by their own submission.

The final phase of the challenge was open for three weeks.
At this stage, the leaderboard was completely private, disallowing participants to see their own scores. 
This choice was made to encourage the evaluation on the validation split rather than the test split and, together with the fact that all ground-truth depths were withheld, severely avoiding any possibility of overfitting over the test set by conducting repeated evaluations on it.

Following the second edition \cite{spencer2023second}, any form of supervision was allowed, in order to provide a more comprehensive overview of the monocular depth estimation field as a whole.
This makes it possible to better study the gap between different techniques and identify possible, future research directions.
In this paper, we report results only for submissions that outperformed the baseline in any pointcloud-/image-based metric on the Overall dataset.

\heading{Dataset}
The challenge takes place based on the \acl{syns} dataset~\cite{Adams2016,Spencer2022}, chosen due to the diversity of scenes and environments. 
A breakdown of images per category and some representative examples are shown in \fig{data:distrib} 
and \fig{data:syns}.
\acl{syns} also provides extremely high-quality dense ground-truth \acs{lidar}, with an average coverage of 78.20\% (including sky regions). 
Given such dense ground-truth, depth boundaries were obtained using Canny edge-detection on the log-depth maps, allowing us to compute additional fine-grained metrics for these challenging regions.
As outlined in~\cite{Spencer2022,spencer2023second}, the images were manually checked to remove dynamic object artifacts.

\heading{Evaluation}
Participants were asked to provide the up-to-scale disparity prediction for each dataset image. 
The evaluation server bilinearly upsampled the predictions to the target resolution and inverted them into depth maps.
Although self-supervised methods trained with stereo pairs and supervised methods using \acs{lidar} or RGB-D data should be capable of predicting metric depth, 
in order to ensure comparisons are as fair as possible, the evaluation aligned any predictions with the ground-truth using the median depth. 
We set a maximum depth threshold of 100 meters. 

\addfig*[!t]{syns}{data:syns}

\heading{Metrics}
Following the first and second editions of the challenge~\cite{Spencer2023,spencer2023second}, we use a mixture of image-/pointcloud-/edge-based metrics.
Image-based metrics are the most common (MAE, RMSE, AbsRel) and are computed using pixel-wise comparisons between the predicted and ground-truth depth map. 
Pointcloud-based metrics~\cite{Ornek2022} (F-Score, IoU, Chamfer distance) instead bring the evaluation in the 3D domain, evaluating the reconstructed pointclouds as a whole.
Among these, we select reconstruction F-Score as the leaderboard ranking metric. 
Finally, edge-based metrics are computed only at depth boundary pixels. 
This includes image-/pointcloud-based metrics and edge accuracy/completion metrics from IBims-1~\cite{Koch2018}.

\section{Challenge Submissions} \label{sec:submit}
We now highlight the technical details for each submission, as provided by the authors themselves. 
Each submission is labeled based on the supervision used, including ground-truth (\textbf{D}), proxy ground-truth (\textbf{D*}), DepthAnything~\cite{Yang2024} pretraining ($\dagger$) and monocular (\textbf{M}) or stereo (\textbf{S}) photometric support frames. Teams are numbered according to rankings.

\subsection*{Baseline -- S}
\emph{%
\begin{tabular}{ll}
    J.\ Spencer & j.spencermartin@surrey.ac.uk \\
    C.\ Russell & chris.russell@oii.ox.ac.uk \\ 
   S.\ Hadfield & s.hadfield@surrey.ac.uk \\
     R.\ Bowden & r.bowden@surrey.ac.uk \\
\end{tabular}
}

\noindent
Challenge organizers' submission from the first edition. 
\\
\heading{Network}
ConvNeXt-B encoder~\cite{Liu2022} with a base Monodepth decoder~\cite{Mayer2016,Godard2017} from~\cite{Spencer2022}.
\\
\heading{Supervision}
Self-supervised with a stereo photometric loss~\cite{Garg2016} and edge-aware disparity smoothness~\cite{Godard2017}.
\\
\heading{Training}
Trained for 30 epochs on \acl{kez} with an image resolution of \shape{192}{640}{}{}.

\subsection*{\teamsec{1}}
\emph{%
\begin{tabular}{ll}
            G. Zhou & zhouguangyuan@bytedance.com \\
            Z. Li & lizhengxin17@mails.ucas.ac.cn \\
            Q. Rao & raoqiang@bytedance.com \\
            Y. Bao & baoyiping@bytedance.com \\
            X. Liu & liuxiao@foxmail.com \\
\end{tabular}
}

\noindent
\heading{Network}
Based on Depth-Anything \cite{Yang2024} with a BEiT384-L backbone, starting from the authors' weights pre-trained on 1.5M labeled images and 62M+ unlabeled images.
\\
\heading{Supervision}
The model is fine-tuned in a supervised manner, with proxy labels derived from stereo images. The final loss function integrates the SILog loss, SSIL loss, Gradient loss, and Random Proposal Normalization (RPNL) loss. 
\\
\heading{Training}
The network was fine-tuned on the CityScapes dataset \cite{Cordts2016Cityscapes}, resizing the input to \shape{384}{768}{}{} resolution, while keeping proxy labels at \shape{1024}{}{}{2048} resolution. Random flipping is used to augment data, the batch size is set to 16 and the learning rate to 0.000161. The fine-tuning is carried out to predict metric depth and early stops at 4 epochs, a strategic choice to prevent overfitting and ensure the model's robustness to new data.

\subsection*{\teamsec{3}}
\emph{%
\begin{tabular}{ll}
            D. Kim & figure317@rgarobot.com \\
            J. Kim & jsk24@rgarobot.com \\
            M. Kim & wiseman218@rgarobot.com \\
\end{tabular}
}

\noindent
\heading{Network}
It uses the Depth Anything \cite{Yang2024} pre-trained model to estimate relative depth, 
accompanied by an auxiliary network to convert it into metric depth. This latter is NAFNet \cite{chen2022simple}, processing the final feature maps and relative depth map predicted by the former model together with the input image.
\\
\heading{Supervision}
Self-supervised loss with two main terms: image reconstruction loss and smoothness loss. The former integrates perceptual loss with photometric loss as used in monodepth2 \cite{Godard2019}, with the former using a pre-trained VGG19 backbone \cite{simonyan2014very}, following a similar approach as in ESRGAN \cite{Wang_2018_ECCV_Workshops}.
\\
\heading{Training}
The train is carried out on \acl{kez} with batch size 8 and learning rate 1$\mathrm{e}{-4}$ for 4 epochs. Only NAFNet is trained, while the Depth Anything model remains frozen.

\subsection*{\teamsec{4}}
\emph{%
\begin{tabular}{ll}
           M.\ Lavreniuk & nick\_93@ukr.net \\
\end{tabular}
}

\noindent
\heading{Network}
The architecture is based on Depth Anything \cite{Yang2024}, incorporating a VIT-L encoder \cite{Dosovitskiy2021} for feature extraction and the ZoeDepth metric bins module \cite{Bhat2023} as a decoder. This module computes per-pixel depth bin centers, which are linearly combined to produce metric depth.
\\
\heading{Supervision}
The models were trained in a supervised manner using ground-truth depth information obtained from various datasets, employing the SILog loss function.
\\
\heading{Training}
The models were trained on both indoor and outdoor data, respectively on the NYUv2 dataset \cite{Silberman2012} with an image size of \shape{392}{518}{}{}, and on KITTI \cite{Geiger2013}, Virtual KITTI 2 \cite{cabon2020vkitti2}, and DIODE outdoor \cite{diode_dataset} with an image size of \shape{518}{1078}{}{}. The batch size was set to 16, the learning rate to 0.000161, and the maximum depth to 10 for indoor scenes. For outdoor scenes, the batch size was set to 1, the learning rate to 0.00002, and the maximum depth to 80. Both models were trained for 5 epochs.

\subsection*{\teamsec{6}}
\emph{%
\begin{tabular}{ll}
R. Li & lirui.david@gmail.com \\
Q. Mao & maoqing@mail.nwpu.edu.cn \\
J. Wu & 18392713997@mail.nwpu.edu.cn \\
Y. Zhu & yuzhu@nwpu.edu.cn \\
J. Sun & sunjinqiu@nwpu.edu.cn \\
Y. Zhang & ynzhang@nwpu.edu.cn \\
\end{tabular}
} 

\noindent
\heading{Network}
An architecture made of two sub-networks.
The first model consists of a pre-trained ViT-large backbone \cite{Dosovitskiy2021} from Depth Anything \cite{Yang2024} and a ZoeDepth decoder \cite{Bhat2023}. The second is Metric3D \cite{yin2023metric3d}, which uses ConvNext-Large \cite{liu2022convnet} backbone and a LeRes decoder \cite{Wei2021CVPR}.
\\
\heading{Supervision} 
The first network is fine-tuned with the KITTI dataset using SILog loss. The second network uses the released pre-trained weights trained by a diverse collection of datasets as detailed in \cite{yin2023metric3d}.
\\
\heading{Training}
The first network is fine-tuned using batch size 16 for 5 epochs. 
At inference, test-time augmentation -- i.e., color jittering and horizontal flipping -- is used to combine the predictions by the two models: the same image is augmented 10 times and processed by the two models, then the predictions are averaged.

\subsection*{\teamsec{7}}
\emph{%
\begin{tabular}{ll}
S. Patni & suraj.patni@cse.iitd.ac.in \\
A. Agarwal & aradhye.agarwal.cs520@cse.iitd.ac.in \\
C. Arora & chetan@cse.iitd.ac.in \\
\end{tabular}
}

\noindent
\heading{Network}
The model is ECoDepth \cite{Suraj2024ecodepth}, which provides effective conditioning for the MDE task to diffusion methods like stable diffusion.
It is based on a Comprehensive Image Detail Embedding (CIDE) module which utilizes ViT embeddings of the image and subsequently transforms them to yield a semantic context vector. These embeddings are used to condition the pre-trained UNet backbone in Stable Diffusion, which produces hierarchical feature maps from its decoder. These are resized to a common dimension and passed to the Upsampling decoder and depth regressor to produce the final depth. 
\\
\heading{Supervision} 
Supervised training using the ground truth depth with SILog
loss as the loss function with variance focus ($\lambda$) 0.85. Ground-truth depth is transformed as $\frac{1}{(1+x)}$. 
\\
\heading{Training}
Trained on NYUv2 \cite{Silberman2012}, KITTI \cite{Geiger2013}, virtual KITTI v2 \cite{cabon2020vkitti2} for 25 epochs, with one-cycle learning rate (min: 3$\mathrm{e}{-5}$, max: 5$\mathrm{e}{-4}$) and batch size 32 on 8$\times$ A100 GPUs.

\subsection*{\teamsec{9}}
\emph{%
\begin{tabular}{ll}
P. Sun & 23s136164@stu.hit.edu.cn \\
K. Jiang & jiangkui@hit.edu.cn \\
G. Wu & gwu@hit.edu.cn \\
J. Liu & hitcslj@hit.edu.cn \\
X. Liu & csxm@hit.edu.cn \\
J. Jiang & jiangjunjun@hit.edu.cn \\
\end{tabular}
}

\noindent
\heading{Network}
It involves the pre-trained Depth Anything encoder and pre-trained CLlP model. The latter is introduced to calculate the similarity between the keywords `indoor' or `outdoor' and features extracted from the input image to route it to two, different instances of Depth Anything specialized on indoor or outdoor scenarios. \\
\heading{Supervision}
Two instances of Depth Anything are fine-tuned on ground-truth labels, respectively from NYUv2 and KITTI for indoor and outdoor environments.
\\
\heading{Training}
The training resolution is \shape{392}{518}{}{} on NYUv2 and \shape{384}{768}{}{} on KITTI. The batch size is 16 and both instances are trained for 5 epochs.

\subsection*{\teamsec{10}}
\emph{%
\begin{tabular}{ll}
         X.\ Zhang & zhangxidan@fujitsu.com \\
         J.\ Wei & weijianing@fujitsu.com \\
         F.\ Wang & wangfangjun@fujitsu.com \\
         Z.\ Tan & zhmtan@fujistu.com \\
\end{tabular}
}

\noindent
\heading{Network}
The depth network is the Depth Anything \cite{Yang2024} pre-trained model -- based on ZoeDepth \cite{Bhat2023} with a DPT\_BEiT\_L384 -- and further fine-tuned. 
\\
\heading{Supervision}
Trained on ground-truth depth, with SILog and Hyperbolic Chamfer Distance losses. 

\heading{Training}
The model is fine-tuned on NYU-v2 \cite{Silberman2012}, 7Scenes \cite{7scene}, SUNRGBD \cite{sunrgbd}, DIODE \cite{diode_dataset}, KITTI \cite{Geiger2013}, DDAD \cite{ddad}, and Argoverse \cite{argoverse} -- without any resizing of the image resolution -- for 20 epochs with batch size 32, a learning rate set to 1.61e-04, and a 0.01 weight decay.

\subsection*{\teamsec{15}}
\emph{%
\begin{tabular}{ll}
         J. Wang & 601533944@qq.com \\
\end{tabular}
}

\noindent
\heading{Network}
Swin encoder \cite{Liu2021b} with skip connections and a decoder with channel-wise self-attention modules.
\\
\heading{Supervision}
Trained with ground truth depths, using a loss consisting of a combination of two L1 losses and an SSIM loss, weighted accordingly.
\\
\heading{Training}
The model was trained on \acl{kez} split using images of size \shape{370}{1224}{}{} for 100 epochs.

\subsection*{\teamsec{16}}
\emph{%
\begin{tabular}{ll}
         A.\ Luginov & a.luginov@pgr.reading.ac.uk \\
         M.\ Shahzad & m.shahzad2@reading.ac.uk \\
\end{tabular}
}

\noindent
\heading{Network}
The depth network is SwiftDepth \cite{swiftdepth}, a compact model with only 6.4M parameters.
\\
\heading{Supervision}
Self-supervised monocular training with the minimum reconstruction loss \cite{Godard2019}, enhanced by offline knowledge distillation from a large MDE model \cite{Yang2024}.
\\
\heading{Training}
The model is trained in parallel on \acl{kez} and a selection of outdoor YouTube videos, similarly to KBR \cite{Spencer2023c}. Both training and prediction are performed with the input resolution of \shape{192}{640}{}{}. The teacher model \cite{Yang2024} is not trained on either these datasets or SYNS-Patches.

\subsection*{\teamsec{19}}
\emph{%
\begin{tabular}{ll}
S. Hosseini & smhh@yorku.ca \\
A. Trajcevski & atrajcev@yorku.ca \\
J. H. Elder & jelder@yorku.ca \\
\end{tabular}
}

\noindent
\heading{Network}
An off-the-shelf semantic segmentation model \cite{wang2023internimage} is used at first to segment the image. Then, the depth of pixels on the ground plane is estimated by predicting the camera angle from the height of the highest pixel on the ground. Then, depth is propagated vertically for pixels above the ground, while the Manhattan frame is estimated with \cite{qian2019ls3d} to identify both Manhattan and non-Manhattan segments in the image and propagate depth along them in 3D space. Finally, the depth map is completed according to heat equations \cite{elder1999edges}, with pixels for which depth has been already estimated imposing forcing conditions, while semantic boundaries and the image frame impose reflection boundary conditions.
\\
\heading{Supervision}
Ground-truth depth is used for training three kernel regression models. \\
\heading{Training}
Three simple statistical models are trained on CityScapes \cite{Cordts2016Cityscapes} and NYUv2 \cite{Silberman2012}: 1) A kernel regression model to estimate ground elevation angle from the vertical image coordinate of the highest observed ground pixel. The ground truth elevation angle is computed by fitting a plane (constrained to have zero roll) to the ground truth ground plane coordinates; 2) A kernel regression model to estimate the depth of ground pixels from their vertical coordinate, conditioned on semantic class; 3) median depth of non-ground pixels in columns directly abutting the bottom of the image frame, conditioned on semantic class. \\

\section{Results} \label{sec:res}

Submitted methods were evaluated on the testing split of \acl{syns}~\cite{Adams2016,Spencer2022}.
Participants were allowed to submit methods without any restriction on the supervision or the predictions by the model, which can be either relative or metric.
Accordingly, to ensure a fair comparison among the methods, the submitted predictions are aligned to ground-truth depths according to median depth scaling.

{ 
\addtbl*[!t]{results}{res:results}
}

\subsection{Quantitative Results}

\tbl{res:results} highlights the results of this third edition of the challenge, with the top-performing techniques, ordered using F-Score performance, achieving notable improvements over the baseline method. A first, noteworthy observation is the widespread adoption of the Depth Anything model \cite{Yang2024}, pre-trained on 62M of images, as the backbone architecture by the leading teams, including ~\team{1}, ~\team{3}, ~\team{4}, ~\team{6}, ~\team{9}, ~\team{10}, and ~\team{16}, demonstrating its effectiveness and versatility. 

Specifically, Team ~\team{1}, which secured the top position on the leaderboard, achieved an F-score of 23.72, outperforming the baseline method by a remarkable $72.9\%$. This represents a significant improvement over the previous state-of-the-art method, DJI\&ZJU, which achieved an F-score of 17.51 in the ``The Second Monocular Depth Estimation Challenge" \cite{spencer2023second}. In particular, Team ~\team{1}'s result shows a $35.5\%$ increase in performance compared to DJI\&ZJU, highlighting the rapid progress made in monocular depth estimation within a relatively short period. This improvement can be also clearly observed in the other metrics considered, both accuracy and error -- notably, achieving the second absolute results on F-Edges, MAE, and RMSE. Their success can be attributed to the fine-tuning of the Depth Anything model on the Cityscapes dataset using a combination of SILog, SSIL, Gradient, and Random Proposal Normalization losses, as well as their strategic choice of fine-tuning for a few epochs to prevent overfitting and ensure robustness to unseen data. 

Team ~\team{3}, in the third place, achieved an F-score of 22.79, outperforming the baseline by $66.1\%$. Their novel approach of augmenting the Depth Anything model, maintained frozen, with an auxiliary network, NAFNet, to convert relative depth predictions into metric depth, combined with self-supervised loss terms, shows the effectiveness of this approach in enhancing depth accuracy. In terms of the F-Edges metric, this method achieves the best result.

Team ~\team{4}, ranking fourth, achieved an F-score of 20.87, surpassing the baseline by 52.1\%. Their approach involved training the Depth Anything model on both indoor and outdoor datasets, adapting image sizes, batch sizes, and learning rates to each scenario, and highlighting the importance of tailoring model parameters to the specific characteristics of the target environment. This strategy notably improves the results in terms of standard 2D error metrics, yielding the lowest MAE, RMSE, and AbsRel. 

Several other teams also surpassed both the baseline method and the previous state-of-the-art from the second edition of the challenge. Team ~\team{6} achieved an F-score of 20.42, outperforming the baseline by 48.8\% by fine-tuning and combining predictions from the Depth Anything model and Metric3D. Team ~\team{7} follows surpassing the baseline using ECoDepth, which conditions Stable Diffusion's UNet backbone with Comprehensive Image Detail Embeddings. Team ~\team{9} and ~\team{10} also achieved notable improvements, with F-score of 17.83 and 17.81, respectively, using specialized model instances and fine-tuning on diverse datasets. 

Finally, the remaining teams outperformed the baseline either on the F-score or any of the other metrics, yet not surpassing the winner of the previous edition. Team \team{15}, with an F-score of 15.92, outperformed the baseline by 16.0\% using a Swin encoder with skip connections and a decoder with channel-wise self-attention modules, while Team \team{16} outperforms the baseline by distilling knowledge from Depth Anything to a lightweight network based on SwiftDepth, further improved using minimal reconstruction loss during training.
Finally, Team ~\team{19} employed an off-the-shelf semantic segmentation model and estimated depth using techniques such as predicting camera angle, propagating depth along Manhattan and non-Manhattan segments, and completing the depth map using heat equations. They achieved an F-score of 11.04, 19.5\% lower than the baseline score of 13.72, yet they obtained 3.22 Acc-Edge, beating the baseline. 

\addfig*[!t]{depth}{res:depth_viz}

\subsection{Qualitative Results}

\fig{res:depth_viz} provides qualitative results for the depth predictions of each submission. A notable trend among the top-performing teams, such as ~\team{1}, ~\team{3}, ~\team{4}, and ~\team{6}, is the adoption of the Depth Anything model as a backbone architecture. While Depth Anything represents the current state-of-the-art in monocular depth estimation, the qualitative results highlight that there are still significant challenges in accurately estimating depth, particularly for thin structures in complex outdoor scenes. This is evident in columns 2, 4, 5, and 6 of \fig{res:depth_viz}, where objects like trees and branches are not well-recovered, despite the impressive quantitative performance of these methods as shown in \tbl{res:results}. Interestingly, Team ~\team{7}, which employs a novel approach called ECoDepth to condition Stable Diffusion's UNet backbone with Comprehensive Image Detail Embeddings, demonstrates a remarkable ability to estimate depth for thin structures. Yet, they are outperformed quantitatively by other methodologies, suggesting that estimating depth in smooth regions may be more challenging than in thin structures.

The qualitative results also reveal some method-specific anomalies. For instance, ~\team{15} exhibits salt-and-pepper noise artifacts, while ~\team{19}'s method, which ranks last, generates overly smooth depth maps that lose important scene objects. These anomalies highlight the importance of developing robust techniques that can handle diverse scene characteristics. Grid-like artifacts are observed in the predictions of top-performers ~\team{1} and ~\team{3}, particularly in regions where the network seems uncertain about depth estimates. This suggests that further improvements in network architecture and training strategies may be necessary to mitigate these artifacts.

The indoor scenario in the last column shows the strong performance of methods like ~\team{1}, ~\team{4}, ~\team{9}, and ~\team{10} in estimating scene structure. This can be attributed to their use of large-scale pre-training, fine-tuning on diverse datasets, and carefully designed loss functions that capture both global and local depth cues.

However, all methods still exhibit over-smoothing issues at depth discontinuities, manifesting as halo effects. While they outperform the baseline in this regard, likely due to their supervised training with ground truth or proxy labels, there remains significant room for improvement.

A notable limitation across all methods is the inability to effectively estimate depth for non-Lambertian surfaces, such as glass or transparent objects. This is evident in the penultimate right column and the first column, corresponding to the windshield. The primary reason for this limitation is the lack of accurate supervision for such surfaces in the training data, highlighting the need for novel techniques and datasets that explicitly address this challenge.

In conclusion, the qualitative results provide valuable insights into the current state of monocular depth estimation methods. While the adoption of large-scale pre-training and carefully designed architectures has led to significant improvements, challenges persist in accurately estimating depth for thin structures, smooth regions, and non-Lambertian surfaces. Addressing these limitations through novel techniques, improved training strategies, and diverse datasets will be crucial for further advancing this field.

\section{Conclusions \& Future Work} \label{sec:conclusion}
This paper has summarized the results for the third edition of \ac{mdec}.
Over the various editions of the challenge, we have seen a drastic improvement in performance, showcasing \ac{mde} -- in particular real-world generalization -- as an exciting and active area of research. 

With the advent of the first foundational models for \ac{mde} during the last months, we observed a diffused use of frameworks such as Depth Anything \cite{Yang2024}. This ignited a major boost to the results submitted by the participants, with a much higher impact compared to the specific kind of supervision chosen for the challenge. 
Nonetheless, as we can appreciate from the qualitative results, any methods still struggle to accurately predict fine structures and discontinuities, hinting that there is still room for improvement despite the massive amount of data used to train Depth Anything.

We hope \ac{mde} will continue to attract new researchers and practitioners to this field and renew our invitation to participate in future editions of the challenge.

\textbf{Acknowledgments.}
This work was partially funded by the EPSRC under grant agreements EP/S016317/1, EP/S016368/1, EP/S016260/1, EP/S035761/1.

{
\small
\bibliographystyle{ieee_fullname}
\bibliography{main}
}

\end{document}

%% file: Preamble/preamble.tex
\usepackage{acronym}
\usepackage{adjustbox}
\usepackage{booktabs}
\usepackage[font=small,labelfont=bf]{caption}
\usepackage{colortbl}
\usepackage{fmtcount}
\usepackage{forloop}
\usepackage{multirow}
\usepackage{stackengine}
\usepackage{suffix}
\usepackage[para]{footmisc}  
\usepackage{svg}
\usepackage{rotating}
\usepackage{pifont}    
\usepackage{stmaryrd}  
\usepackage{tabularx}
\usepackage[textsize=scriptsize]{todonotes}
\usepackage{xfrac}
\usepackage{xr}
\usepackage{xspace}

\presetkeys{todonotes}{inline}{}  

\makeatletter
\newcommand*{\addFileDependency}[1]{
  \typeout{(#1)}
  \@addtofilelist{#1}
  \IfFileExists{#1}{}{\typeout{No file #1.}}
}
\makeatother

\definecolor{ForestGreen}{HTML}{228b22}  

\DeclareRobustCommand{\nbd}{\nobreakdash-} 

\newcommand{\pnorm}[1]{L\ensuremath{_#1}}

\newcommand{\heading}[1]{\noindent \textbf{\small #1.}}

\newcommand{\addfig}[3][htbp]{%
\begin{figure}[#1]

\centering
\input{Figures/#3}
\label{fig:#3}

\end{figure}
}

\WithSuffix\newcommand\addfig*[3][htbp]{%
\begin{figure*}[#1]

\centering
\input{Figures/#3}
\label{fig:#3}

\end{figure*}
}

\newcommand{\addtbl}[3][htbp]{%
\begin{table}[#1]

\scriptsize
\addtolength{\tabcolsep}{-0.2em}
\renewcommand{\arraystretch}{1.1}
\centering
\input{Tables/#3}
\label{tbl:#3}

\end{table}
}

\WithSuffix\newcommand\addtbl*[3][htbp]{%
\begin{table*}[#1]

\scriptsize
\addtolength{\tabcolsep}{-0.2em}
\renewcommand{\arraystretch}{1.1}
\centering
\input{Tables/#3}
\label{tbl:#3}

\end{table*}
}

\def\eg{\emph{e.g}.~}

\newcommand{\fig}[1]{Figure~\ref{fig:#1}}
\newcommand{\tbl}[1]{Table~\ref{tbl:#1}}

\newcommand{\acromath}[3]{\acrodef{#1}[\(#2\)]{#3}} 
\newcommand{\myac}[1]{\text{\acs{#1}}}  

\newcommand{\shape}[4]{\ensuremath{  
#1 \times #2
\ifthenelse {\equal{#3}{}} {} {\times #3}
\ifthenelse {\equal{#4}{}} {} {\times #4}
}}

\newcommand{\sca}[1]{\ensuremath{#1}}  						   
\newcommand{\vct}[1]{\ensuremath{\textbf{\MakeLowercase{#1}}}} 
\newcommand{\mat}[1]{\ensuremath{\textbf{\MakeUppercase{#1}}}} 
\newcommand{\set}[1]{\ensuremath{\MakeUppercase{#1}}} 	       

\newcommand{\thr}{\delta}



%% file: Preamble/acronyms.tex
\acrodef{ar}     [AR]             {Augmented Reality}
\acrodef{araug}  [AR-Aug]         {aspect ratio augmentation}

\acrodef{cnn}    [CNN]            {Convolutional Neural Network}

\acrodef{d25}    [$\delta_{.25}$] {$\delta < 1.25$}
\acrodef{dnn}    [DNN]            {Deep Neural Network}

\acrodef{fscore} [F]              {F-Score}

\acrodef{imu}    [IMU]            {Inertial Measurement Unit}

\acrodef{kbr}    [KBR]            {Kick Back \& Relax}

\acrodef{lidar}  [LiDAR]          {Light Detection and Ranging}

\acrodef{mae}    [MAE]            {Mean Absolute Error}
\acrodef{mde}    [MDE]            {monocular depth estimation}
\acrodef{mdec}   [MDEC]           {Monocular Depth Estimation Challenge}
\acrodef{ml}     [ML]             {Machine Learning}

\acrodef{rel}    [Rel]            {Absolute Relative Error}

\acrodef{sfm}    [SfM]            {Structure-from-Motion}
\acrodef{slam}   [SLAM]           {Simultaneous Localization and Mapping}
\acrodef{sota}   [SotA]           {State-of-the-Art}
\acrodef{ssl}    [SSL]            {self-supervised learning}

\acrodef{vo}     [VO]             {Visual Odometry}

\acrodef{crf}      [NeWCRFs] {NeWCRFs}

\acrodef{ddad}     [DDAD]    {DDAD}
\acrodef{dii}      [DIODE]   {DIODE Indoors}
\acrodef{dio}      [DIODE]   {DIODE Outdoors}
\acrodef{dpt-beit} [DPT]     {DPT-BEiT}
\acrodef{dpt-vit}  [DPT]     {DPT-ViT}

\acrodef{kb}       [KB]      {Kitti Benchmark}
\acrodef{ke}       [KE]      {Kitti Eigen}
\acrodef{keb}      [KEB]     {Kitti Eigen{\nbd}Benchmark}
\acrodef{kez}      [KEZ]     {Kitti Eigen{\nbd}Zhou}

\acrodef{mc}       [MC]      {Mannequin Challenge}
\acrodef{midas}    [MiDaS]   {MiDaS}

\acrodef{nyud}     [NYUD]    {NYUD{\nbd}v2}

\acrodef{sintel}   [Sintel]  {Sintel}
\acrodef{ssmde}    [SS-MDE]  {SS{\nbd}MDE}
\acrodef{stv}      [STV]     {SlowTV}
\acrodef{syns}     [SYNS]    {SYNS{\nbd}Patches}

\acrodef{tum}      [TUM]     {TUM{\nbd}RGBD}

%% file: Preamble/symbols.tex
\acromath{pix}        { \vct{p} } {}

\acromath{point-gt}   { \vct{q} } {}
\acromath{point-pred} { \hat{\myac{point-gt}} } {}
 
\acromath{depth-gt}   { \sca{y} } {}
\acromath{depth-pred} { \hat{\myac{depth-gt}} } {}
 
\acromath{cloud-gt}   { \set{Q} } {}
\acromath{cloud-pred} { \hat{\myac{cloud-gt}} } {}
 
\acromath{edges-gt}   { \mat{M} } {}
\acromath{edges-pred} { \hat{\acs{edges-gt}} } {}
 
\acromath{thr-3d}     { \sca{\thr} } {}

